\documentclass[conference]{IEEEtran}
\usepackage{siunitx}

\usepackage{booktabs}
\usepackage{booktabs}
\usepackage{tikz}
\usetikzlibrary{positioning}
\tikzset{out/.style={draw, fill=orange!20, rectangle, minimum height=1.2em, minimum width=2em, align=center}}

\IEEEoverridecommandlockouts
\usepackage{hyperref}

\usepackage{cite}
\usepackage{amsmath,amssymb,amsfonts}
\usepackage{algorithmic}
\usepackage{graphicx}
\usepackage{textcomp}
\usepackage{xcolor}
\usepackage{url}
\usepackage{tikz}
\usepackage{booktabs}
\usepackage{cuted}
\usepackage{siunitx}
\usepackage{capt-of}
\usepackage{multirow}  
\usepackage{graphicx}  
\usetikzlibrary{arrows.meta,positioning,calc}
\usetikzlibrary{arrows.meta,positioning}
\usepackage[ruled,linesnumbered]{algorithm2e}
\def\BibTeX{{\rm B\kern-.05em{\sc i\kern-.025em b}\kern-.08em
    T\kern-.1667em\lower.7ex\hbox{E}\kern-.125emX}}
\begin{document}

\title{Context-Aware Causal Gaze Forecasting for Human–Vehicle Interaction During In-Cabin Tracking Dropouts\\}
\title{Context-Aware Causal Gaze Forecasting for
Human--Vehicle Interaction During In-Cabin Tracking Dropouts}

\author{
\IEEEauthorblockN{Shabnam Shabani}
\IEEEauthorblockA{
\textit{Department of Computer Science}\\
\textit{Western University}\\
London, ON, Canada\\
sshaban7@uwo.ca
}
\and
\IEEEauthorblockN{Ghazal Farhani}
\IEEEauthorblockA{
\textit{National Research Council Canada}\\
London, ON, Canada\\
ghazal.farhani@nrc-cnrc.gc.ca
}
}

\maketitle

\maketitle

\begin{abstract}
Dashboard-mounted gaze trackers often lose sight of the driver's eyes
during large head rotations, such as shoulder checks, mirror glances,
and intersection scanning. These maneuvers are also among the moments
when information about the driver's visual attention is most useful.
Offline gap-filling methods may reconstruct a missing interval using
observations from both sides of the gap, but an online driver-monitoring
system cannot rely on measurements that have not yet occurred. We
therefore formulate \emph{causal gaze recovery}: forecasting unavailable
gaze at time $t$ without access to target-tracker gaze at $t$ or any
later time.

We introduce the \textbf{Causal Context-Gated Forecaster (CCGF)}, which
encodes a 60-frame pre-dropout history of gaze and head pose and combines
it with DINOv3 scene features. A learned reliability gate controls the contribution of the
history and scene representations as the dropout progresses. We
evaluate two scene-availability conditions: \emph{Live}, in which the
scene representation continues to update during tracker loss, and
\emph{Frozen}, in which the final pre-dropout representation is used
throughout the missing interval.

We evaluate CCGF on $2{,}047$ eligible, naturally occurring GazeSense
\texttt{head\_lost} events drawn from $10.5$\,h of naturalistic driving
by ten drivers. Across the complete recordings, \texttt{head\_lost}
accounts for $8.5\%$ of GazeSense recording time. Synchronized gaze
coordinates from a head-mounted Neon tracker provide supervision and
evaluation targets but are never used as model inputs. Under
leave-one-driver-out evaluation, CCGF achieves a mean per-driver median
error of $175.7$\,px ($10.5^\circ$) with Live scene updates, a $33\%$
reduction relative to history-only causal forecasting. With Frozen
scene input, the error increases to $210.8$\,px ($12.9^\circ$),
indicating that scene observations acquired during the dropout provide
useful predictive information. We will release the dataset, evaluation
protocol, and causal baselines.
\end{abstract}

\begin{IEEEkeywords}
driver monitoring, causal gaze forecasting, gaze tracking, tracking dropout,
driver attention, naturalistic driving.
\end{IEEEkeywords}
\begin{figure}[t]
\centering

\includegraphics[width=\columnwidth, height=\columnwidth]{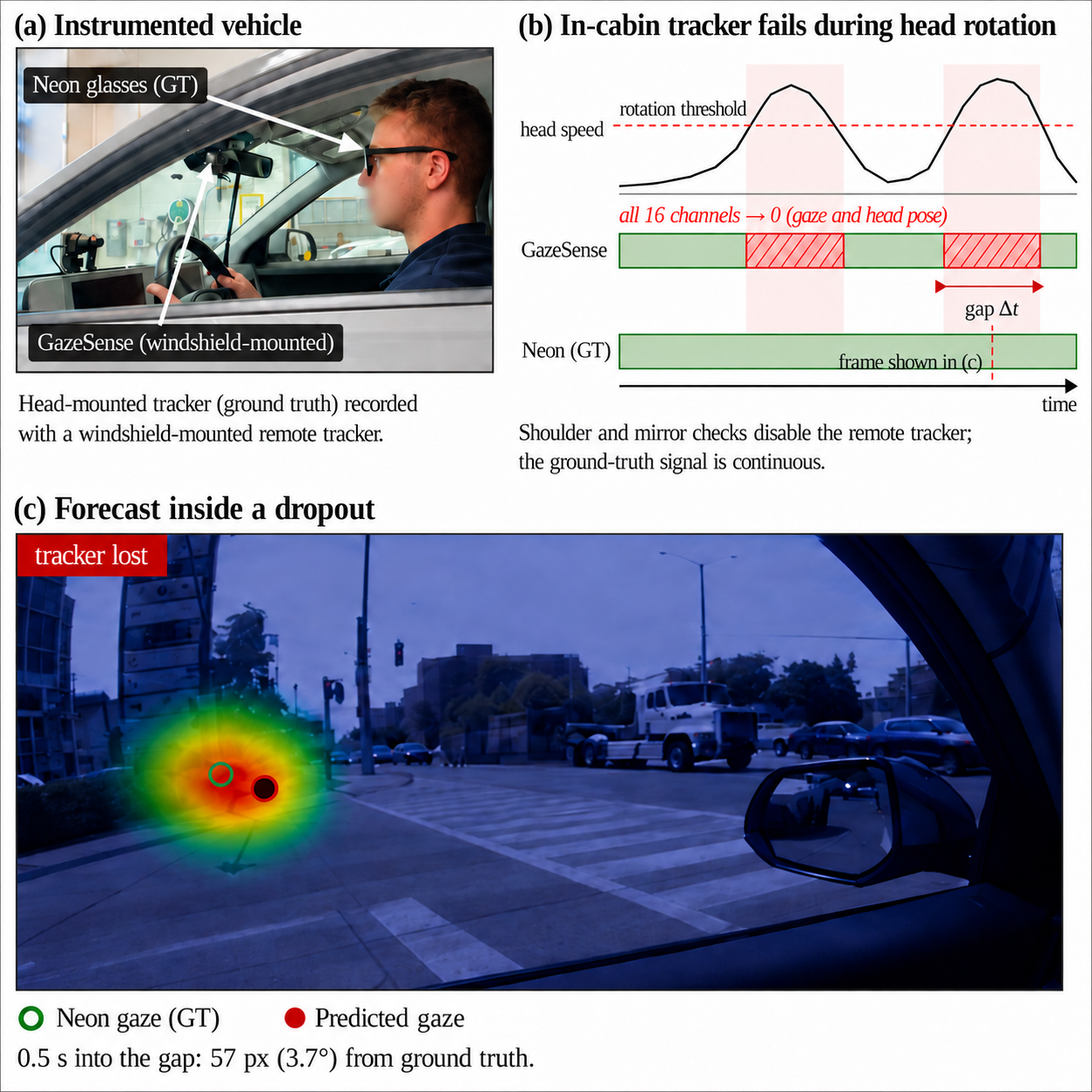} 
\caption{\textbf{Recovering driver gaze through in-cabin tracking dropouts.}
\textbf{(a)}~Drivers are recorded with a head-mounted Pupil Labs Neon
eye tracker, which provides reference gaze during large head rotations,
and a windshield-mounted GazeSense remote tracker representing the
in-cabin sensing stream.
\textbf{(b)}~Shoulder checks, mirror glances, and other large head
rotations can cause GazeSense to enter a \texttt{head\_lost} state,
while the head-mounted reference remains available throughout the
dropout $\Delta t$.
\textbf{(c)}~During tracker loss, CCGF forecasts gaze causally from
pre-dropout GazeSense history and the currently available scene, without
using future observations. In the illustrated example, the prediction is
57\,px ($3.7^\circ$) from the Neon reference 0.5\,s into the dropout.}
\label{fig:teaser}

    \label{fig:teaser}
\end{figure}

\section{Introduction}

Driver-monitoring systems are increasingly integrated into advanced
driver-assistance systems (ADAS) to infer whether a driver is attentive,
aware of surrounding traffic, and ready to act when needed
\cite{sae2021taxonomy}. Gaze is among the most informative behavioral
cues because it directly indicates where visual attention is allocated,
and has been used for maneuver anticipation \cite{brain4cars,
kung_lookinginsideout, wu_gaze_intention}, weather and situation
awareness \cite{shabani2022analysis, biswas_sa, farhani2025weather}, and
takeover-readiness assessment \cite{gaze2movement}.

These applications assume gaze is continuously observable. In real
driving it is not. Dashboard-mounted trackers require a clear view of
the driver's eyes, yet ordinary driving involves large head rotations
during mirror checks, shoulder checks, lane changes, and intersection
scanning \cite{headpose_rgbd, gazezone_rgbd}. During such movements the
eyes leave the tracker's field of view and the signal is lost, precisely
when the driver's visual attention is most safety-relevant. In our
naturalistic recordings the dashboard tracker is unavailable for 8.5\%
of driving time, concentrated in moments such as blind-spot checks
before a turn or merge.

We ask whether gaze can be estimated during these events \emph{while
the dropout is still occurring}. This differs from conventional
missing-data reconstruction: time-series imputation methods use
observations on both sides of a gap \cite{brits, saits, csdi,
imputeformer, sssd}, whereas an online driver-monitoring system cannot
access observations that have not yet occurred. The operational problem
is \emph{strictly causal forecasting}: at every instant during a
dropout, gaze must be estimated from information available up to that
instant only.

Studying this problem requires ground truth during the very intervals
in which the in-cabin tracker has failed. We therefore record drivers
simultaneously with a dashboard-mounted GazeSense tracker
\cite{gazesense} and head-mounted Pupil Labs Neon eye-tracking glasses
\cite{pupilneon}. GazeSense provides the deployable in-cabin signal
together with its naturally occurring failures, while Neon continues to
measure gaze through large head rotations. After synchronizing and
cross-calibrating both systems into a common scene coordinate frame,
Neon \emph{gaze} measurements serve only as supervision and evaluation
targets and are never supplied as predictive inputs. The outward-facing
Neon scene camera, by contrast, provides the synchronized visual stream,
standing in for the fixed vehicle camera a deployed system would use.

From these recordings we construct and release a dataset of real,
rotation-induced gaze dropouts from ten drivers, preserving the timing,
duration, and head-motion patterns of genuine tracker failures rather
than synthetically masked gaps.

As a reference solution we introduce \textbf{CCGF} (Causal
Context-Aware Gaze Forecasting), which combines the valid pre-dropout
gaze and head-motion history with contemporaneous visual context encoded
by a frozen DINOv3 backbone \cite{dinov3}, without access to
post-dropout gaze. We compare CCGF against classical causal
extrapolation, general-purpose learned forecasters, and non-causal
imputers; the latter use future observations and are reported only as
offline references. Because a fixed vehicle camera would not follow the
driver's head, we evaluate two regimes: a live-scene regime that uses
the egocentric view during the gap, giving an optimistic lower bound on
error, and a frozen-scene regime with no new scene image during the gap,
giving a worst-case upper bound.

The main contributions of this work are:
The main contributions of this work are:
\begin{itemize}
    \item We formulate recovery over genuine in-cabin gaze-tracker
    dropouts as a \emph{strictly causal forecasting} problem, distinct
    from offline bidirectional imputation.

    \item We introduce a synchronized naturalistic-driving dataset
    collected from ten drivers, containing genuine remote-tracker
    dropouts paired with head-mounted reference gaze. This enables
    evaluation on real rather than synthetically generated tracker
    failures.\footnote{An
\href{https://drive.google.com/drive/folders/1cHy3AVKOTmLiG_71IIVPwBTnw74xHeyD}
{anonymized example recording} is available for review. The complete
dataset, code, and ethics approval details will be made publicly
available upon acceptance.}

    \item We develop CCGF and evaluate it alongside classical causal
    extrapolation, learned causal forecasting, and non-causal imputation
    methods. This comparison isolates the contributions of pre-dropout
    gaze and head-pose history and live visual context during tracker
    loss.
\end{itemize}

\section{Related Work}

\subsection{Driver Gaze Estimation from Observable Eyes}

Most driver-gaze estimation methods assume that the driver's face and
eyes are visible at prediction time. DGAZE~\cite{dgaze} maps gaze to the
road scene from a cropped eye image, head pose, face location, and
related geometric cues, reporting a localization error of 186.89 pixels
on a $1920\times1080$ road view. DashGaze~\cite{dashgaze} addresses
continuous gaze estimation during naturalistic on-road driving from
driver, face, eye, and facial-landmark inputs, and reports variation
across camera positions and cases in which the face or one eye is only
partially visible. Errors across such studies are not directly
comparable because datasets, target representations, resolutions, and
metrics differ, but they indicate that gaze localization in driving
scenes remains a hard problem even with direct access to the eye region.

Robustness to head pose is a longstanding difficulty in appearance-based
gaze estimation. Gaze360~\cite{gaze360} and ETH-XGaze~\cite{ethxgaze}
extend benchmarks to wide pose and appearance variation, and
3DGazeNet~\cite{threedgazenet} improves generalization by explicitly
modeling three-dimensional eye geometry. Other work handles degraded
visibility through richer appearance models, additional sensors, or
multiple cameras. All of these improve \emph{estimation} while some
visual evidence of the eyes remains available; none address the case in
which the tracker has entered a genuine \texttt{head\_lost} interval and
returns no valid gaze measurement at all.

\subsection{Missing Data and Causal Forecasting}

A tracker dropout can also be viewed as a missing-data problem. Deep
time-series imputation methods reconstruct unobserved values from the
surrounding sequence: BRITS~\cite{brits} uses bidirectional recurrent
dynamics, SAITS~\cite{saits} self-attention, GP-VAE~\cite{gpvae} a
Gaussian-process latent prior, and CSDI~\cite{csdi} conditional
score-based diffusion. These methods exploit observations that occur
after the missing interval, which an online driver-monitoring system
cannot access while the failure is in progress. We therefore report
BRITS, SAITS, and GP-VAE as non-causal offline references rather than
deployable predictors.

For the causal setting we compare against general-purpose forecasting
architectures: DLinear~\cite{dlinear}, a decomposition-based linear
model; TimesNet~\cite{timesnet}, which models temporal variation through
a two-dimensional multi-period representation; and SegRNN~\cite{segrnn},
which uses segment-wise recurrence with parallel multi-step output. We
also evaluate CSDI under a past-only forecasting protocol. All causal
baselines receive only the pre-dropout GazeSense-derived history and
never see future GazeSense observations or reference gaze.

Our problem sits at the intersection of these two settings. Unlike gaze
estimators, the predictor has no current eye-based observation during
the interval of interest; unlike bidirectional imputation, it cannot use
post-recovery measurements. We study strictly causal gaze forecasting
during naturally occurring remote-tracker failure.

\section{Methodology}
\label{sec:method}
\subsection{Dataset and Problem Setup}
\label{subsec:data}

We collected on-road recordings from ten drivers, comprising
$10.5$h of naturalistic driving. A windshield-mounted GazeSense remote tracker~\cite{gazesense} provides the deployable gaze and head-pose
stream, while a head-mounted Pupil Labs Neon system~\cite{pupilneon}
provides independent reference gaze and an egocentric scene video.
GazeSense \texttt{head\_lost} intervals account for 8.5\% of recorded
driving time and form 2,047 naturally occurring tracker-failure events
before benchmark filtering.

The two streams are synchronized at 30\,Hz and expressed in a common
$1600{\times}1200$ Neon scene coordinate system. Per-recording
geometric alignment uses a Kabsch-initialized rotation followed by an
affine ray-space correction. Timestamp offset and linear drift are
estimated independently for each recording. Neon gaze inside GazeSense
failure intervals is excluded from calibration and is never used as a
predictor input.

\noindent\textbf{Dropout events and eligibility.}
A real dropout is defined as a maximal contiguous GazeSense interval
labeled \texttt{head\_lost}. Each event is paired with the preceding
$L=60$ frames ($\approx2$\,s) of causal context, providing recent
pre-failure gaze and head-motion history.

We retain failures lasting 3--300 frames (0.1--10\,s). The lower bound
removes isolated missing samples, while the upper bound excludes rare
prolonged interruptions outside the temporary tracker failures targeted
here. At least 70\% of failure frames must contain valid Neon reference
gaze to ensure sufficient ground-truth coverage. We additionally require
a valid GazeSense observation immediately before failure onset and at
least 50\% GazeSense availability within the preceding context.
The latter criterion retains meaningful observed history without
preferentially removing clustered failures, which commonly contain
short periods of prior tracker loss. Temporal continuity is required
throughout the context--failure interval.

For scene-conditioned evaluation, at least 90\% of failure frames must
have a causal scene representation no older than 100\,ms
($\approx3$ frames at 30\,Hz), ensuring predominantly contemporaneous
visual information while tolerating occasional dropped frames.
Frozen additionally requires a valid pre-dropout scene representation.
These criteria yield 2{,}047 eligible naturally occurring dropout events.

For duration-stratified analysis only, events are grouped into short
($<0.5$\,s), medium ($0.5$--$1.5$\,s), long ($1.5$--$3$\,s), and
very-long ($>3$\,s) intervals. The boundaries were chosen from the
empirical dropout-duration distribution to separate sub-second failures,
the main body of intermediate-duration events, and the relatively rare
long-duration tail; at 30\,Hz they correspond to approximately 15, 45,
and 90 frames, respectively.

\noindent\textbf{Input representation.}
GazeSense is represented by a 16-D deployable state comprising 3-D gaze
direction, head-pose quaternion and translation, projected 2-D gaze,
tracker availability, timing information, and head angular speed.
Only observations available up to tracker-loss onset are used as gaze
and head-pose inputs. Missing GazeSense measurements during
\texttt{head\_lost} intervals are neither interpolated nor supplied to
the predictor; tracker availability and elapsed failure time remain
explicitly represented.

Each causally available scene frame is encoded by a frozen DINOv3
ViT-B/16 encoder~\cite{dinov3}, producing a 768-D visual representation.
Scene features provide predictor-side visual context, whereas Neon gaze
is used exclusively as reference supervision and for evaluation.

\subsection{Causal Forecasting Formulation}
\label{subsec:formulation}

Let $t_0$ denote the final valid GazeSense frame before tracker loss,
and let $t_0+1,\ldots,t_1$ denote the failure interval. Let
$\mathbf{x}_t$ denote the deployable GazeSense state,
$\mathbf{s}_t$ the causal scene representation, and
$\mathbf{g}_t$ the Neon reference gaze. Under the \emph{Live} setting,
CCGF estimates

\begin{equation}
p\!\left(
\mathbf{g}_t
\mid
\mathbf{x}_{t_0-L+1:t_0},
\mathbf{s}_{t_0-L+1:t}
\right),
\qquad
t_0<t\leq t_1 .
\label{eq:causal}
\end{equation}

This formulation enforces the online constraint directly: GazeSense
observations terminate at $t_0$, while scene information is available
only up to the current prediction time $t$. No future GazeSense
measurements, future scene observations, or Neon gaze targets are
available to the predictor, and the eventual duration of the failure is
unknown at inference time.

We evaluate two primary scene conditions. In \emph{Live}, the scene
pathway continues to receive causally updated visual representations
throughout tracker loss. In \emph{Frozen}, the final valid pre-dropout
scene state is held fixed after $t_0$, so no new visual information is
incorporated during the failure. A separate \emph{scene-free} variant
disables the visual pathway entirely and is used only as an ablation to
quantify the contribution of scene information.

\subsection{CCGF Architecture}
\label{subsec:model}

CCGF contains separate causal pathways for pre-dropout driver history
and visual context. The 60-frame GazeSense context is projected to
128 dimensions and encoded by a two-layer unidirectional LSTM. Once
tracking is lost, missing GazeSense measurements are not recursively
supplied to this pathway. Instead, the final pre-dropout history state
is conditioned on elapsed failure time using a sinusoidal positional
encoding,

\begin{equation}
\widetilde{\mathbf h}^{H}_t
=
\mathbf h^{H}_{t_0}
+
\mathbf W_{\Delta}\mathrm{PE}(\tau_t),
\label{eq:history}
\end{equation}

where $\tau_t$ denotes the elapsed time since tracker loss.

The 768-D DINOv3 representation is projected to 64 dimensions and
processed by a one-layer unidirectional scene LSTM. Under Live input,
the scene state continues to update as new causal scene observations
arrive. Under Frozen input, the final pre-dropout representation is
repeated. Invalid scene observations do not advance the scene pathway.

The history and scene pathways are combined through a learned scalar
gate,

\begin{align}
\alpha_t &=
\sigma\!\left(
f_g[
\widetilde{\mathbf h}^{H}_t
\Vert
\mathbf h^{S}_t]
\right), \\
\mathbf h^{F}_t &=
\alpha_t\widetilde{\mathbf h}^{H}_t+
(1-\alpha_t)\mathbf W_S\mathbf h^{S}_t ,
\label{eq:fusion}
\end{align}

where $\alpha_t=1$ corresponds to history-only prediction and
$\alpha_t=0$ to scene-only prediction. If no valid visual context is
available, the model falls back to the history pathway.

The fused representation is decoded to a $24{\times}32$ spatial
probability distribution $\mathbf p_t$. The predicted gaze is its
center of mass,

\begin{equation}
\hat{\mathbf g}_t
=
\sum_k p_{t,k}\mathbf c_k ,
\label{eq:heatmap_mean}
\end{equation}

where $\mathbf c_k$ denotes the normalized center of grid cell $k$.
The spatial spread of the distribution provides a corresponding
uncertainty estimate and is also used by the probabilistic training
objective.


\begin{figure*}[t]
    \centering
    \includegraphics[width=0.78\textwidth]{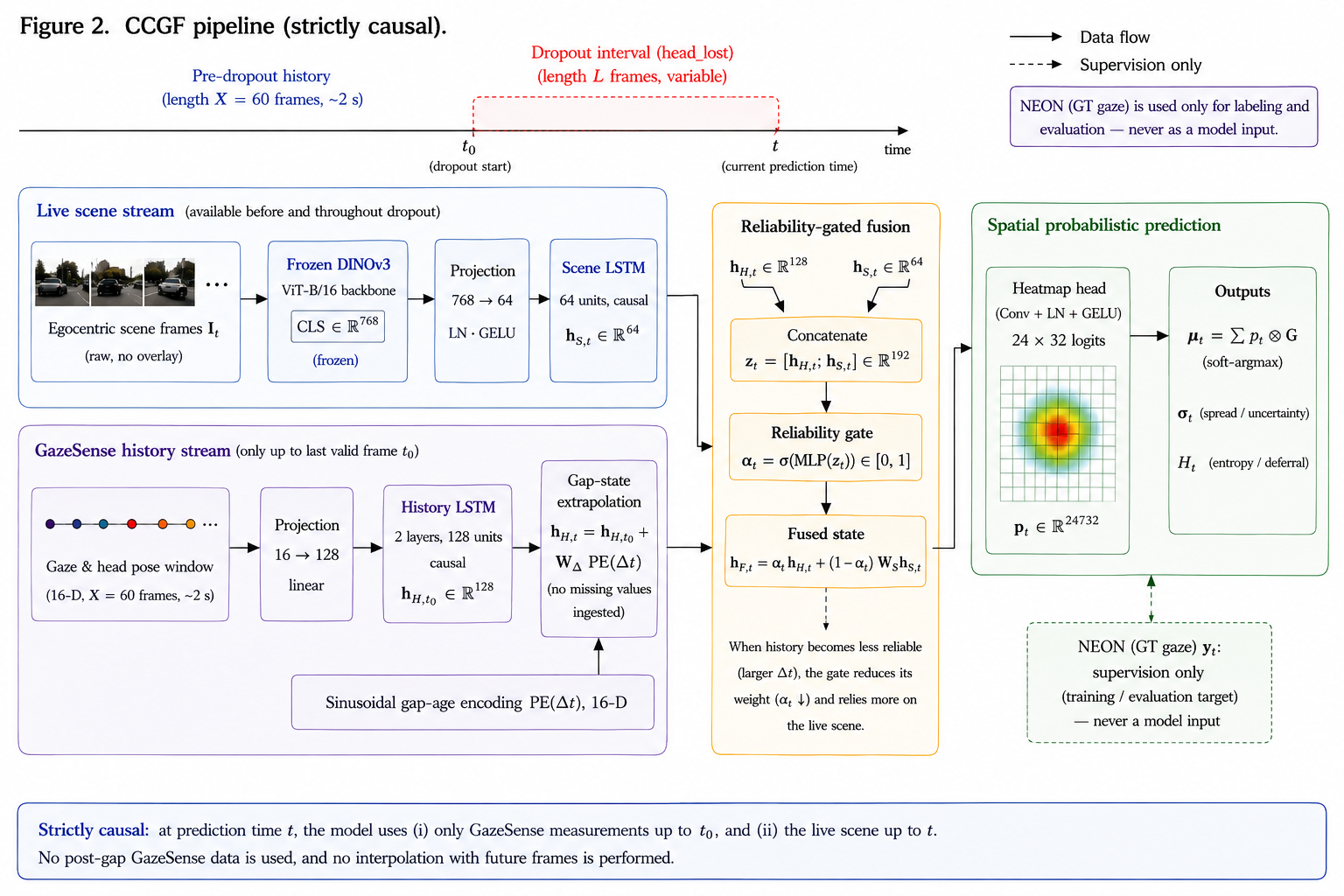}
    \caption{\textbf{CCGF causal forecasting pipeline.}
    The history pathway encodes the valid 60-frame pre-dropout
    GazeSense context and is subsequently conditioned on elapsed failure
    time without receiving missing GazeSense measurements. In parallel,
    the scene pathway processes causally available DINOv3 visual
    representations. A learned scalar gate combines the two pathways,
    and a spatial prediction head forecasts gaze during tracker loss.
    Neon gaze is used only for supervision and evaluation.}
    \label{fig:arch}
\end{figure*}


\begin{algorithm}[t]
\caption{CCGF forecasting during tracker loss}
\label{alg:ccgf}
\small

\SetKwInOut{Input}{Input}
\SetKwInOut{Output}{Output}

\Input{pre-dropout GazeSense context; causal scene stream}
\Output{gaze forecast $\hat{\mathbf g}_t$}

Encode GazeSense history into $\mathbf h^H_{t_0}$\;
Encode pre-dropout scene context into $\mathbf h^S_{t_0}$\;

\For{$t=t_0+1,\ldots$ while GazeSense is unavailable}{

    condition $\mathbf h^H_{t_0}$ on elapsed failure time\;

    update scene state using the current causal scene representation
    (Live) or the final pre-dropout representation (Frozen)\;

    compute fusion weight $\alpha_t$\;

    fuse history and scene states using Eq.~\ref{eq:fusion}\;

    predict spatial distribution $\mathbf p_t$\;

    $\hat{\mathbf g}_t\leftarrow\sum_k p_{t,k}\mathbf c_k$\;

    \textbf{emit} $\hat{\mathbf g}_t$\;
}
\end{algorithm}

\subsection{Training and Evaluation}
\label{subsec:training}
\noindent\textbf{Training.}
Training combines naturally occurring \texttt{head\_lost} events with
synthetic failures sampled from continuously tracked intervals. A real
event is sampled with probability 0.3; otherwise, a synthetic failure
is generated with duration sampled from the empirical distribution of
real events in the corresponding training partition. The GazeSense
payload is masked over the synthetic failure while the causal scene
stream is preserved. Synthetic failures are used only for training;
validation and test sets contain exclusively naturally occurring
tracker failures.

CCGF is optimized using
\begin{equation}
\mathcal{L}
=
\mathcal{L}_{\mathrm{NLL}}
+
0.5\,\mathcal{L}_{\mathrm{hm}}
+
0.1\,\mathcal{L}_{\mathrm{ctx}},
\label{eq:loss}
\end{equation}
where $\mathcal{L}_{\mathrm{NLL}}$ supervises the probabilistic gaze
prediction, $\mathcal{L}_{\mathrm{hm}}$ supervises the spatial heatmap,
and $\mathcal{L}_{\mathrm{ctx}}$ provides auxiliary supervision on
valid pre-dropout context frames. The weights 0.5 and 0.1 are selected
through nested driver-disjoint hyperparameter tuning using only the
development drivers; the outer test driver is never involved in their
selection.

AdamW is used with learning rate $10^{-3}$, weight decay $10^{-4}$,
batch size 16, dropout 0.1, and gradient clipping at 1.0. Training runs
for at most 30 epochs, with checkpoint and stopping decisions made only
from inner-validation performance.

\noindent\textbf{Nested driver-disjoint protocol.}
Generalization is evaluated using nested leave-one-driver-out (LODO)
cross-validation. In each outer fold, one driver is held out entirely
for testing and all remaining drivers form the development set.
Hyperparameter and training-duration selection are performed within
this development set using an inner driver-disjoint LODO procedure:
each development driver serves once as the inner validation driver,
with the remaining drivers used for training. Candidate configurations
are ranked by the mean driver-level median pixel error across the inner
validation folds.

After inner model selection, the selected configuration is retrained
from scratch on the complete development set and evaluated on the
held-out outer driver. Multiple recordings from the same participant
always remain in the same partition. Final learned-model evaluation is
repeated across three independent training seeds.

\noindent\textbf{Baselines.}
We compare CCGF with both simple motion references and learned
time-series methods. Zero-order hold (ZOH) repeats the final valid
projected GazeSense gaze throughout tracker loss. Constant velocity
(CV) fits a linear trajectory to the five most recent valid gaze
observations and falls back to ZOH when fewer than two observations are
available.

DLinear, TimesNet, SegRNN, and CSDI are evaluated as strict causal
temporal forecasters. Each receives the same 60-frame pre-dropout
GazeSense-derived history, with no scene features, Neon gaze, or future
GazeSense observations. Input normalization is estimated from the
training fold only. Because these models require a fixed prediction
length, they forecast 30 frames (1\,s) at a time. If tracker loss
continues, the model is rolled forward using its own predicted 2-D gaze
as pseudo-history while unavailable GazeSense channels remain missing.
The true terminal dropout duration is never supplied to the model and is
used only by the offline evaluator to terminate prediction at tracker
recovery.

BRITS, SAITS, and GP-VAE are included separately as non-causal
imputation references. Their input consists of projected GazeSense gaze
with the failure interval marked missing and additionally includes
60 post-recovery frames. These methods can therefore exploit future
observations to reconstruct the missing interval and represent an easier
offline setting. 

\noindent\textbf{Metrics and aggregation.}
Evaluation is restricted to naturally occurring failure frames with
valid Neon reference gaze. The primary metric is Euclidean pixel error
in the $1600{\times}1200$ scene coordinate system, with angular error
reported as a complementary measure. Because error distributions are
heavy-tailed and recordings contain different numbers of failure
frames, errors are first summarized using the median within each
held-out driver. The primary reported result is the mean of these ten
driver-level medians.

\section{Results}
\label{sec:results}

we evaluate all results on naturally occurring GazeSense
\texttt{head\_lost} events under the driver-disjoint LODO protocol
described in Sec.~\ref{subsec:training}. For each held-out driver, we
compute the median framewise prediction error over eligible dropout
frames and report the mean across the ten test drivers. Learned-model
results are aggregated across three independent training seeds. Angular
error is reported as well.

\subsection{Overall Causal Forecasting Performance}
\label{subsec:causal}

Table~\ref{tab:baselines} compares CCGF with causal alternatives during
real tracker loss: simple motion extrapolation, learned time-series
forecasters, and, as non-causal offline references, bidirectional
imputers.

The motion baselines perform poorly over real dropout intervals.
Zero-order hold, which repeats the last valid gaze estimate throughout
the gap, yields $334.2$\,px, and constant-velocity extrapolation yields
$359.1$\,px. CCGF under the Live scene condition achieves $175.7$\,px
($10.52^\circ$), a reduction of $47.4\%$ and $51.1\%$, respectively.

The learned causal forecasters improve on motion extrapolation but
remain well above CCGF: DLinear reaches $255.8$\,px, TimesNet
$235.4$\,px, SegRNN $279.5$\,px, and CSDI $249.3$\,px, corresponding to
reductions of $31.3\%$, $25.4\%$, $37.1\%$, and $29.5\%$ for CCGF-Live.
Under the more conservative Frozen condition, CCGF reaches $210.8$\,px
($12.91^\circ$), still below every causal baseline in
Table~\ref{tab:baselines}.

All causal methods are subject to the same no-future-information
constraint, and none receives Neon gaze or future GazeSense observations
as input. The temporal baselines see the 60-frame pre-dropout GazeSense
history but no scene features, whereas CCGF additionally uses the
causally available scene pathway. Table~\ref{tab:baselines} therefore
compares complete forecasting systems under strict causality; the
specific contribution of scene context is isolated in
Sec.~\ref{subsec:ablation}.

BRITS, SAITS, and GP-VAE are included as non-causal references. They
may use observations after the missing interval and therefore solve an
easier offline imputation problem than the online forecasting problem
considered here; because they do not use the scene pathway, we treat
them as reference points rather than modality-matched competitors.
Their errors nevertheless remain above CCGF under both scene
conditions. This is a property of the failure mode rather than of the
imputers: a rotation-induced dropout begins and ends with the head
near the forward position, so post-recovery frames resemble
pre-dropout frames, and interpolating between them carries little
information about gaze during the excursion itself. Future
observations are thus not only unavailable online but largely
uninformative for this class of dropout.

\begin{table}[t]
\centering
\small
\caption{Comparison with causal forecasting baselines and non-causal
imputation references on real tracker dropouts. Learned-model results
are aggregated across ten LODO test folds and three independent
training seeds.}
\label{tab:baselines}

\begin{tabular}{p{4.6cm}cc}
\toprule
\textbf{Method} &
\textbf{Error (px)} &
\textbf{Angular ($^\circ$)} \\
\midrule

\multicolumn{3}{l}{\textit{Causal forecasting}} \\

Zero-order hold        & 334.2 & 22.20 \\
Constant velocity      & 359.1 & 23.00 \\
DLinear                & 255.8 & 15.64 \\
TimesNet               & 235.4 & 14.80 \\
SegRNN                 & 279.5 & 16.96 \\
CSDI                   & 249.3 & 15.31 \\

\textbf{CCGF (ours), Live} &
\textbf{175.7} &
\textbf{10.52} \\

\textbf{CCGF (ours), Frozen} &
\textbf{210.8} &
\textbf{12.91} \\

\midrule

\multicolumn{3}{l}{\textit{Non-causal imputation references}} \\

BRITS                  & 243.6 & 15.05 \\
SAITS                  & 233.4 & 14.38 \\
GP-VAE                 & 257.1 & 15.87 \\

\bottomrule
\end{tabular}
\end{table}

\subsection{Ablation: History, Scene Context, and Gated Fusion}
\label{subsec:ablation}

We next isolate the contributions of pre-dropout gaze--head history,
visual context, spatial prediction, and reliability-aware fusion.
Table~\ref{tab:ablation} summarizes the resulting ablation.

\begin{table*}[t]
\centering
\caption{Ablation of scene conditioning and reliability-gated fusion.
Inference time is measured at batch size one and includes DINOv3 scene
feature extraction where applicable.}
\label{tab:ablation}

\begin{tabular}{lccccc}
\toprule
Variant &
Live (px / $^\circ$) &
Frozen (px / $^\circ$) &
None (px / $^\circ$) &
Params &
Inference (ms) $\downarrow$ \\
\midrule

Causal LSTM (gaze $+$ head pose) &
--- &
--- &
263.9 / 16.30 &
269k &
0.78 \\

$+$ scene, concatenation &
176.3 / 10.62 &
244.3 / 15.05 &
\textbf{249.6 / 15.36} &
335k &
10.35 \\

$+$ scene, spatial-heatmap head &
182.2 / 10.95 &
245.9 / 15.11 &
252.5 / 15.59 &
492k &
11.10 \\

\textbf{CCGF: reliability-gated heatmap fusion} &
\textbf{175.7 / 10.52} &
\textbf{210.8 / 12.91} &
258.4 / 15.93 &
505k &
11.38 \\

\bottomrule
\end{tabular}
\end{table*}

\subsubsection{\textbf{Scene Context and Reliability-Aware Fusion}}
\label{subsec:ablation}
With no scene pathway, the causal gaze--head LSTM reaches
$263.9$\,px. Providing continuously updated scene observations reduces
the error substantially: simple scene concatenation reaches
$176.3$\,px, while Live CCGF reaches $175.7$\,px. The small
$0.6$\,px difference between these variants shows that much of the
Live-condition gain comes from access to current visual context itself,
rather than from the gating mechanism alone.

The role of reliability-aware fusion becomes clearer when scene
information becomes stale. Repeating the final pre-dropout scene
embedding throughout tracker loss increases simple concatenation from
$176.3$ to $244.3$\,px and the ungated spatial-heatmap variant from
$182.2$ to $245.9$\,px. In contrast, CCGF reaches $210.8$\,px in
the same Frozen condition, improving over simple concatenation by
$33.5$\,px.

For CCGF, removing within-dropout scene updates increases error from
$175.7$\,px ($10.52^\circ$) to $210.8$\,px ($12.91^\circ$), a loss of
$35.1$\,px and $2.39^\circ$. Frozen CCGF nevertheless remains $10.5\%$
below TimesNet ($235.4$\,px, $14.8^\circ$), the strongest causal
baseline in Table~\ref{tab:baselines}, indicating that the final
pre-dropout scene embedding retains predictive value after tracker
loss.

When scene information is removed entirely, CCGF reaches
$258.4$\,px ($15.93^\circ$), close to the $263.9$\,px
history-only LSTM. Together, these results indicate that visual context
provides the main improvement over temporal history alone, while
reliability-aware fusion becomes particularly useful when that context
becomes stale.

\begin{figure}[t]
    \centering
    \includegraphics[width=0.8\columnwidth]{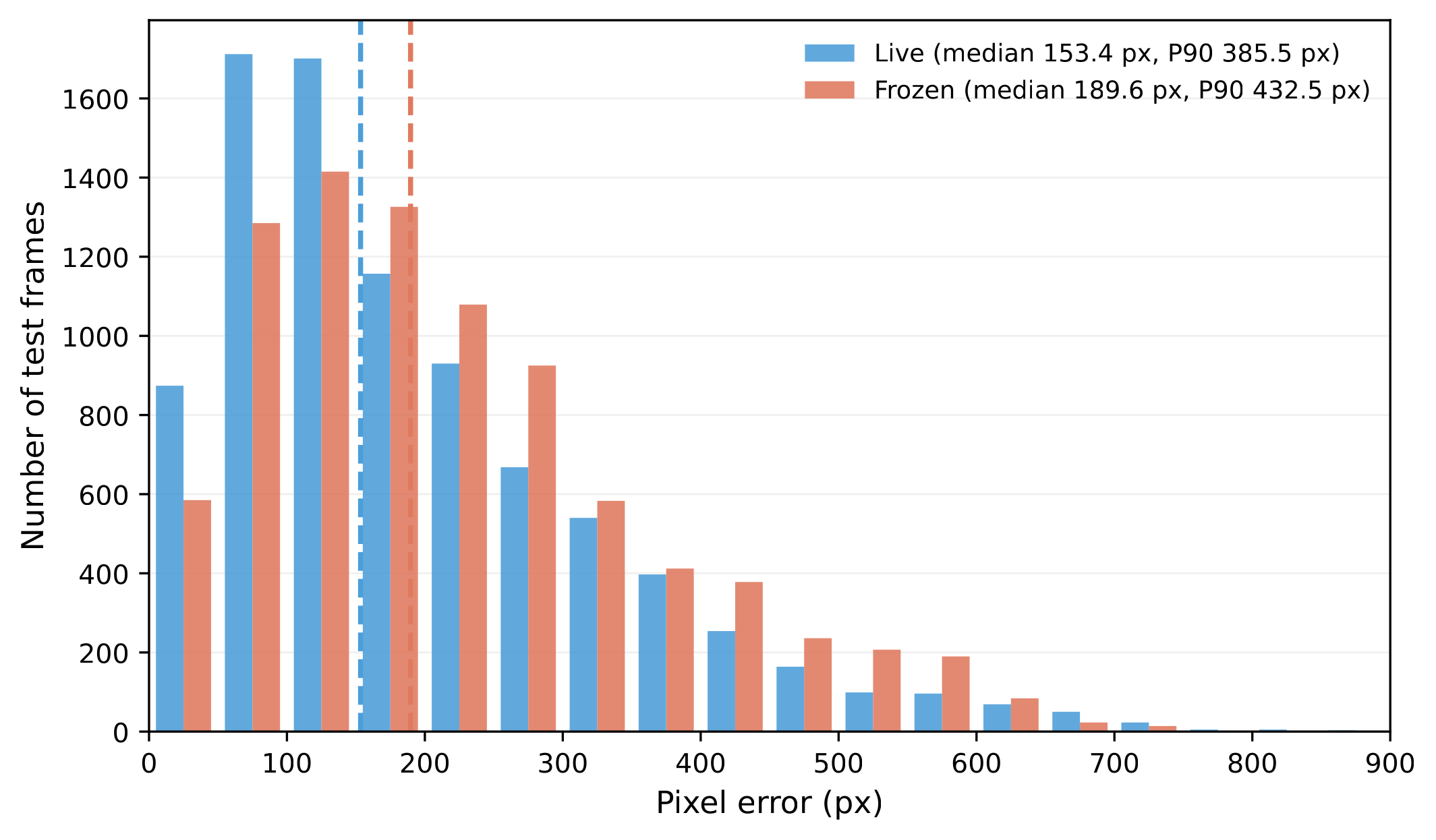}
\caption{\textbf{Illustrative matched Live--Frozen comparison for the
held-out driver 7.}
Pixel-error distributions are shown for the same 8,747 LODO test
frames. Live uses causally updated egocentric scene observations,
whereas Frozen repeats the final pre-dropout scene embedding throughout
tracker loss. In this representative example, median error increases
from 153.4\,px to 189.6\,px and P90 from 385.5\,px to 432.5\,px.
The figure illustrates the effect on matched frames; aggregate
Live--Frozen comparisons are based on the complete ten-driver
evaluation in Table~\ref{tab:ablation}.}
    \label{fig:innara_live_frozen_hist}
\end{figure}

To illustrate how Live vs Frozen scene can affect prediction errors on exactly
the same frames, Fig.~\ref{fig:innara_live_frozen_hist} shows a matched
Live--Frozen example for the held-out driver. For this example, the same 8,747 LODO test frames are evaluated under
both conditions. Median error increases from $153.4$\,px with Live
scene updates to $189.6$\,px when the final pre-dropout scene embedding
is repeated throughout the failure interval. The upper tail also shifts,
with P90 increasing from $385.5$ to $432.5$\,px.

This single-driver example is consistent with the aggregate trend in
Table~\ref{tab:ablation}, but quantitative conclusions are based on the
ten-driver LODO results rather than on this selected example, and complements the
ten-driver aggregate rather than replacing it.

\subsubsection{\textbf{Performance Across Dropout Duration}}
\label{subsec:duration}

Using the predefined dropout-duration bins, performance under the Live
scene condition remains relatively stable over the durations represented
in the dataset. The mean per-driver median error is $165.6$\,px for
short dropouts, $176.7$\,px for medium dropouts, and $180.5$\,px for
long dropouts. The very-long category reaches $176.2$\,px.

Under the Frozen condition, the corresponding errors are $189.4$\,px,
$215.1$\,px, $211.2$\,px, and $199.8$\,px for short, medium, long, and
very-long dropouts, respectively. Frozen therefore produces higher
error in every duration group, with the largest Live--Frozen
differences observed for medium and long failures.

For Live, error increases by only $14.9$\,px (approximately $9\%$)
from short to long dropouts, consistent with the model continuing to
receive causally current scene observations as the last valid GazeSense
measurement becomes older. In contrast, Frozen receives no new visual
evidence after failure onset and therefore relies entirely on the
pre-dropout scene state. Only 61 events fall into the very-long
category, so estimates for this group are interpreted cautiously.

\subsection{Qualitative Analysis}
\label{subsec:qualitative}

\begin{figure}[t]
    \centering
    \includegraphics[width=\columnwidth]{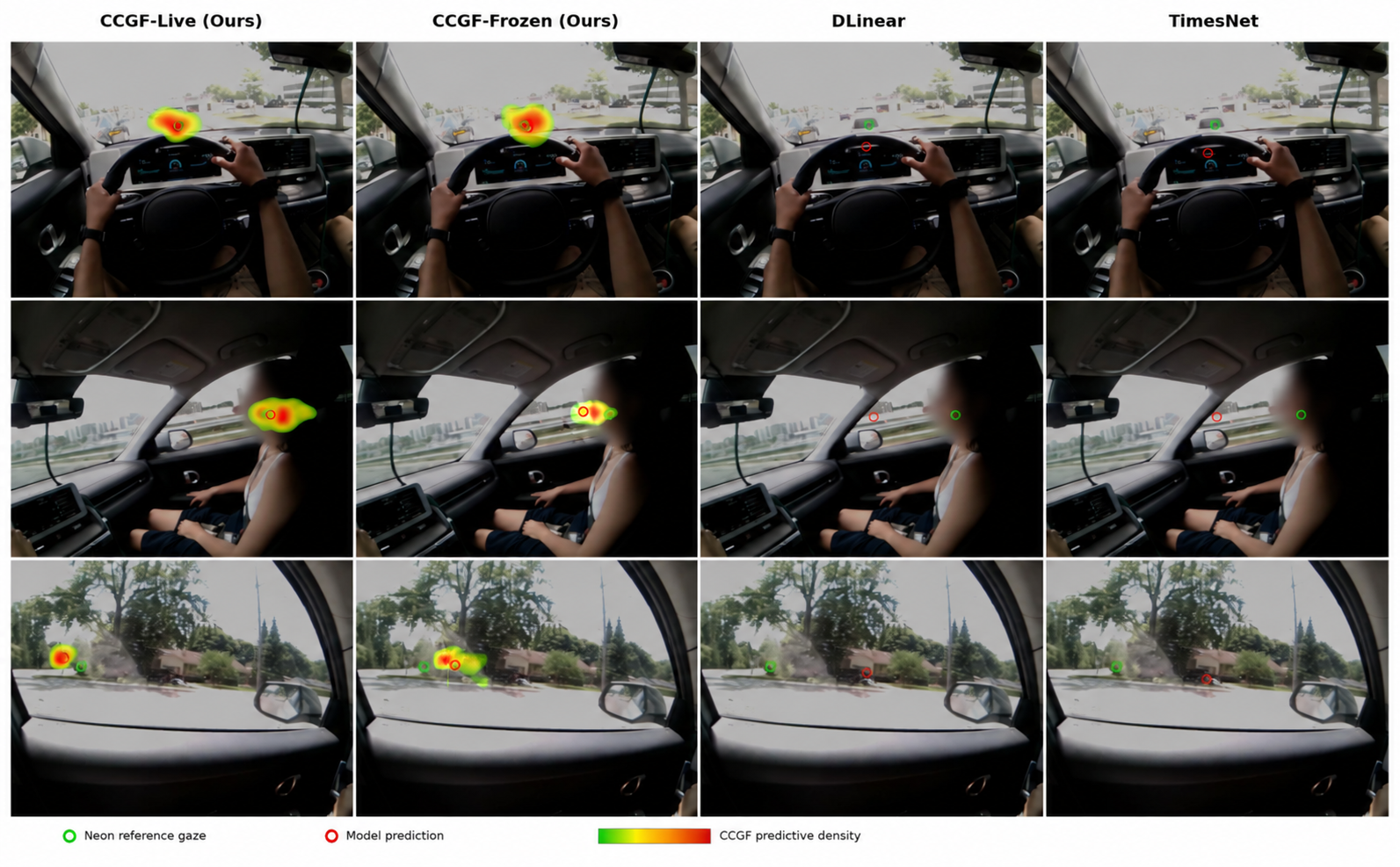}
    \caption{\textbf{Selected qualitative examples during real GazeSense
    tracking failures.}
    Columns compare CCGF with causally updated egocentric scene context
    (Live), CCGF with the final pre-dropout scene representation held
    fixed during tracker loss (Frozen), DLinear, and TimesNet on matched
    frames. Green circles denote Neon reference gaze, red circles denote
    model predictions, and the colored CCGF heatmaps visualize the
    predicted spatial gaze distribution. The examples span
    forward-facing and large-head-rotation conditions. Quantitative
    comparisons are reported over the complete driver-disjoint benchmark
    rather than from these selected examples.}
    \label{fig:qualitative_gaze_prediction}
\end{figure}

Figure~\ref{fig:qualitative_gaze_prediction} shows selected examples
from naturally occurring GazeSense tracker dropouts. The same frames
are compared across CCGF-Live, CCGF-Frozen, DLinear, and TimesNet,
allowing the effect of scene availability and temporal-only forecasting
to be visualized under matched conditions.

These examples are intended to illustrate the spatial behavior of the
predictions rather than to provide additional quantitative evidence.
All comparative conclusions are based on the complete driver-disjoint
evaluation reported in Tables~\ref{tab:baselines}
and~\ref{tab:ablation}; the matched Live--Frozen error distributions
are shown separately in Fig.~\ref{fig:innara_live_frozen_hist}.

\subsection{Runtime and Online Feasibility}
\label{subsec:efficiency}

In the benchmark configuration, CCGF requires approximately
$11.38$\,ms for batch-1 inference, including DINOv3 scene feature
extraction and the forecasting model. This remains below the
$33.3$\,ms frame period of a 30\,Hz input stream.

The DINOv3 backbone is frozen and therefore requires only a forward pass
at deployment. Neon gaze is not required online and is used only as the
independent supervision and evaluation reference. The reported latency
reflects the research implementation rather than a fully optimized
streaming system.

\subsection{Limitations and Future Work}
\label{subsec:limitations}
Several limitations define the scope of the present findings.
First, the Live scene stream comes from the head-mounted Neon camera,
whose viewpoint rotates with the driver's head and may therefore encode
head-motion cues in addition to scene semantics. The Live result should
thus be read as an optimistic case. In the Frozen condition, by
contrast, the scene input stops updating once tracking is lost and is
held at the last frame observed while the head was forward-facing, so
the model receives no visual information about the regions to the left
or right of the vehicle that the driver turns toward. The Frozen result
is therefore a conservative case. Together, the two conditions bracket
the error that can be expected from a deployed system; the exact
mapping from this platform to fixed vehicle-mounted cameras is left for
future work.

Second, the current framework uses only GazeSense-derived driver state
and visual scene context. Vehicle signals such as steering angle,
vehicle yaw and pitch, acceleration, and turn-signal state could provide
additional causal information, particularly during lane changes, turns,
and intersection approaches. Integrating such telemetry with
fixed-camera visual context is a natural next step.

Third, the dataset contains ten drivers. Driver-disjoint LODO
evaluation and three independent training seeds provide a strict test
of generalization and reduce sensitivity to a favorable optimization
run, but larger and more diverse cohorts are needed to assess robustness
across drivers, vehicles, road environments, and sensor configurations.

\section{Conclusion}
\label{sec:conclusion}

This work formulates naturally occurring failures of remote in-cabin
gaze tracking as a strictly causal forecasting problem rather than as
an offline gap-filling problem. Such failures frequently occur during
challenging viewing conditions, including large driver head rotations,
when the remote tracker can no longer maintain a reliable gaze
estimate. CCGF combines valid pre-dropout GazeSense gaze and head-pose
history with causally available scene observations while excluding
future GazeSense measurements and head-mounted Neon gaze from predictor
inputs. Neon gaze is used only as an independent training and
evaluation reference.

Across driver-disjoint LODO evaluation, CCGF achieves a mean
per-driver median error of $175.7$\,px ($10.52^\circ$) when causally
updated egocentric scene observations remain available and
$210.8$\,px ($12.91^\circ$) when the final pre-dropout scene state is
held fixed throughout tracker loss.

Under the Live condition, CCGF reduces pixel error by $47.4\%$
relative to zero-order hold, $51.1\%$ relative to constant velocity,
$31.3\%$ relative to DLinear, $25.4\%$ relative to TimesNet,
$37.1\%$ relative to SegRNN, and $29.5\%$ relative to CSDI.
Even under the Frozen condition, CCGF reduces pixel error by $36.9\%$
relative to zero-order hold and by $10.5\%$ relative to TimesNet, the
strongest learned causal baseline.

The ablation results further show that access to visual context accounts
for much of the improvement over history-only forecasting, while
reliability-aware fusion becomes particularly useful when scene
information is no longer updated. At the same time, the head-mounted
nature of the Live scene stream means that its performance should be
interpreted as an optimistic egocentric condition rather than as a
direct estimate of fixed-camera deployment accuracy.

The central finding is therefore deliberately narrower than exact gaze
reconstruction: a temporary loss of remote gaze tracking does not
necessarily imply a complete loss of information about the driver's
attended region. Even under strict causal constraints, useful gaze
information can be forecast through naturally occurring tracker
failures, including challenging periods involving substantial head
motion. Extending this framework to fixed vehicle-mounted cameras,
vehicle telemetry, and larger driver cohorts is the next step toward
translating this result into a practical driver-monitoring system.

\bibliographystyle{ieeetr} 
\bibliography{references}
\end{document}